\documentclass[10pt,twocolumn]{article}

\usepackage[margin=1in, top=1in]{geometry}

\usepackage{fontspec}

\usepackage{graphicx}
\usepackage{tikz}

\usepackage{xcolor}
\usepackage{fancyhdr}
\usepackage{titlesec}
\usepackage{tcolorbox}
\usepackage{titling}
\tcbuselibrary{skins}

\usepackage{amsmath}
\usepackage{amssymb}

\usepackage{subcaption}
\usepackage{booktabs}
\usepackage{multirow}

\usepackage{algorithm}
\usepackage{algorithmic}

\usepackage[numbers]{natbib}
\usepackage{cleveref}

\usepackage{newfloat}
\usepackage{listings}

\newfontface\titlefont
{Alfabet-Bold.otf} 

\newfontface\emphfont
{Alfabet-Regular.otf}

\newfontface\footerfont
{Alfabet-Light.otf}

\definecolor{qc_blue}{HTML}{2A2AEA}
\definecolor{qc_darkblue}{HTML}{020B3F}

\titleformat{\title}
  {\titlefont\LARGE\color{qc_blue}}
  {}{0pt}{}

\titleformat{\section}
  {\titlefont\Large\color{qc_darkblue}}
  {\thesection}{1em}{}
\titlespacing*{\section}{0em}{1em}{0.6em}

\titleformat{\subsection}
  {\titlefont\large\color{qc_darkblue}}
  {\thesubsection}{1em}{}

\titleformat{\subsubsection}
  {\titlefont\normalsize\color{qc_darkblue}}
  {\thesubsubsection}{1em}{}

\newtcolorbox{titlebox}{
  enhanced,
  colback=white,
  boxrule=0pt,
  opacityback=0,
  opacityframe=0,
  width=0.95\textwidth,
  center
}

\newcommand{\contactinfo}[1]{\def\thecontactinfo{#1}} \contactinfo{}
\contactinfo{}

\fancypagestyle{titlepage}{
  \fancyhf{}
  \fancyfoot[L]{%
    \footerfont\footnotesize
    * Equal Contribution \\
    Qualcomm AI Research is an initiative of Qualcomm Technologies, Inc.
  }
  \fancyfoot[R]{%
    \footerfont\footnotesize\thepage
  }
  
}

\DeclareCaptionStyle{ruled}{
  labelfont=normalfont,
  labelsep=colon,
  strut=off
}

\floatstyle{ruled}
\newfloat{listing}{tb}{lst}{}
\floatname{listing}{Listing}

\title{Surfing the Activation Wave:
\\ Analyzing the Information and Geometry of LLM Hidden States for Speculative Decoding}

\author{
  Raghavv Goel*, Sudhanshu Agrawal*$^\psi$, Mukul Gagrani, Risheek Garrepalli, Mingu Lee
}

\contactinfo{
  Qualcomm AI Research and $^\psi$ Stanford University\\
  \{raghgoel,rgarrepa,mgagrani,mingul\}@qti.qualcomm.com, sudhagra@stanford.edu
}

\date{}

\begin{document}
% \thispagestyle{titlepage}

% % ============================================================
% % QUALCOMM RADIAL GRAPHIC
% % ============================================================
% \begin{tikzpicture}[remember picture,overlay]
%   \node[anchor=north east, inner sep=0pt] at
%     ([xshift=5cm,yshift=5.5cm] current page.north east)
%   {%
%     \includesvg[
%       width=10cm,
%       angle=-75,
%       origin=c,
%       inkscapelatex=false
%     ]{logo/qc_radial-01_rgb_blue}%
%   };
% \end{tikzpicture}

% % ============================================================
% % QUALCOMM AI RESEARCH LOGO
% % ============================================================
% \begin{figure}[t]
%   \vspace*{-1cm}
%   \hspace*{-0.6cm}
%   \includegraphics[width=4.0cm]{logo/qc_air.eps}
%   \vspace*{-0.5cm}
% \end{figure}

% % ============================================================
% % TITLE
% % ============================================================
% \begin{titlebox}
%   {\titlefont \Huge \bfseries \color{qc_darkblue} \thetitle \par}
%   \vspace{1em}

%   % ===========================
%   % AUTHORS
%   % ===========================
%   \makeatletter
%   {\titlefont \color{qc_blue} \@author \par}
%   \vspace{0.25em}
%   {\bfseries \emphfont \@contactinfo \par}
%   \makeatother
%   \vspace{0.5em}

% \end{titlebox}
\thispagestyle{titlepage} % ============================================================ % FULL-WIDTH QUALCOMM TITLE AREA % ============================================================ 
\twocolumn[{ 

% Qualcomm radial graphic 
\begin{tikzpicture}[remember picture,overlay]   
\node[anchor=north east, inner sep=0pt] at      ([xshift=5cm,yshift=5.5cm]current page.north east) 
    % {\includesvg[
    %     width=10cm, 
    %     angle=-75, 
    %     origin=c, 
    %     inkscapelatex=false ]
    %     {logo/qc_radial-01_rgb_blue}}; 
{
    \includegraphics[ width=10cm, angle=-75, origin=c ]{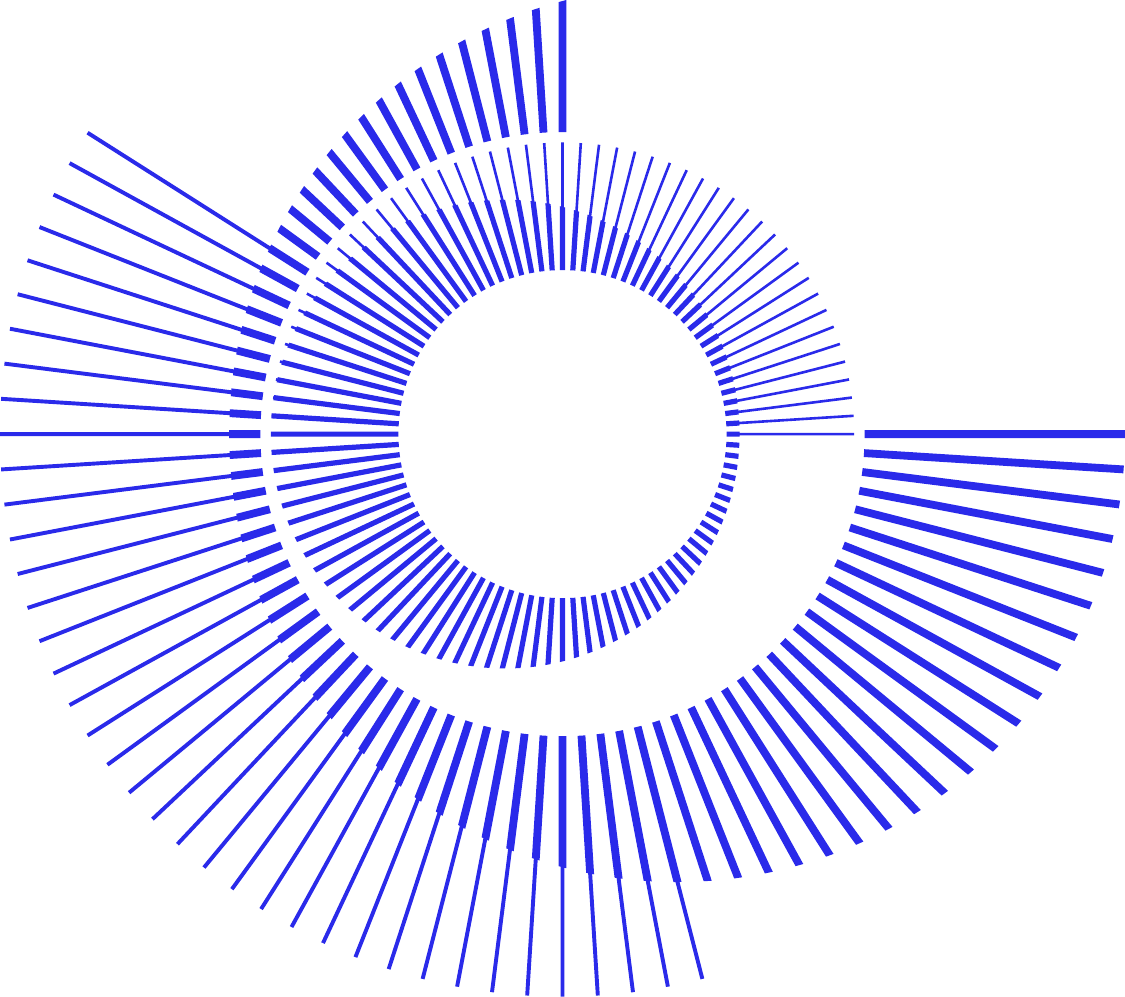} 
};    
\end{tikzpicture} 
% Qualcomm AI Research logo 
\begin{flushleft} 
\includegraphics[width=4.0cm]{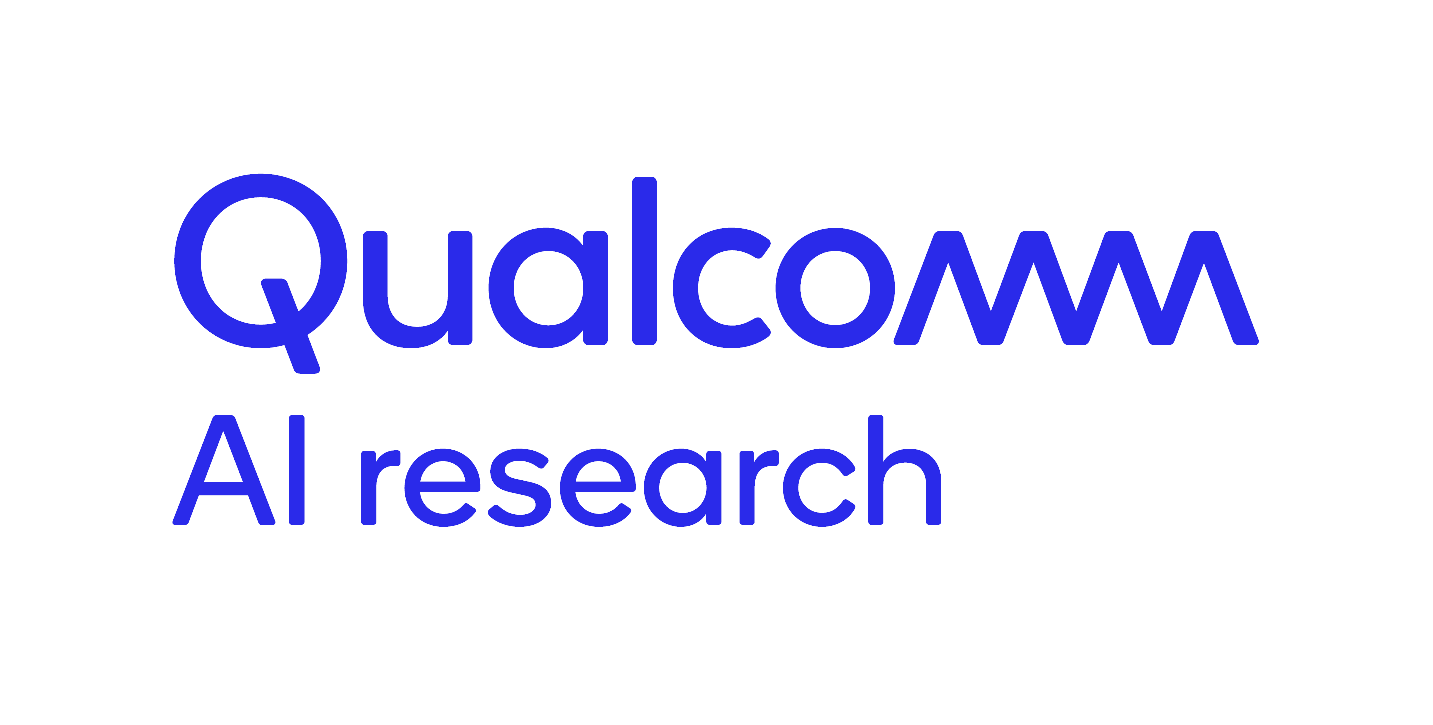} 
\end{flushleft} \vspace{0.3cm} 
% Paper title 
\begin{titlebox} 
{
\titlefont 
\Huge 
\color{qc_darkblue} 
\thetitle 
\par 
} 

\vspace{1em} 
% Authors 
\makeatletter 

{\titlefont 
\color{qc_blue} 
\theauthor 
\par } 

\vspace{0.25em} 
{
\emphfont 
\thecontactinfo 
\par } 

\makeatother 

\end{titlebox} 

\vspace{0.6em} 

}]

% ============================================================ % 

  % ===========================
  % ABSTRACT
  % ===========================
  \begin{abstract}

% Decoder hidden states from different layers in large language models (LLMs) may provide complementary information, but current speculative decoding pipelines typically select a small set of layers from the target model using fixed, position-based rules. The problem of selecting a set of layers from a model is combinatorial, making exhaustive ablations impractical for typical LLMs, even those with moderate depth. In this work, we present a principled framework for multi-layer hidden-state selection using two complementary signals: matrix-entropy-based informativeness and centered-kernel-alignment (CKA) diversity. We analyze per-layer Shannon entropy, total entropy under concatenation, and pairwise CKA transitions, and show that hidden-state geometry forms clear depth-wise compartments. Motivated by these observations, we propose entropy and CKA-based selection algorithms that prefer informative yet non-redundant layer sets. As a component of these algorithms, we additionally introduce token-entropy filtering, where hidden states of low predictive-entropy tokens are removed to sharpen selection statistics. In EAGLE3-style speculative decoding, our selected sets outperform fixed baseline choices, including the default $[2, N//2, N-3]$ rule, while providing a clearer explanation of why some defaults can remain competitive. These results provide a practical recipe for multi-layer hidden-state selection for speculative decoding, insights into the geometry and information in LLM decoder states, and a foundation for extending similar analysis to related applications.

Decoder hidden states from different layers of large language models (LLMs) often encode complementary information. However, existing speculative decoding methods typically rely on a small set of target-model layers chosen using fixed, position-based heuristics. Selecting an optimal subset of layers is inherently a combinatorial problem, making exhaustive evaluation infeasible even for moderately deep models. In this work, we present a principled framework for multi-layer hidden-state selection for speculative decoding based on two complementary criteria: informativeness, measured through matrix-entropy analysis, and diversity, quantified using centered kernel alignment (CKA). We investigate layer-wise Shannon entropy, aggregate entropy under hidden-state concatenation, and pairwise CKA dynamics across depth, revealing distinct geometric and informational compartments within decoder representations. Motivated by these findings, we propose entropy and CKA-based layer selection algorithms that favor hidden-state sets that are both highly informative and minimally redundant. As part of this framework, we introduce token-entropy filtering, which removes hidden states associated with low predictive-entropy tokens, thereby improving the quality of layer-selection statistics. In EAGLE3-style speculative decoding, our selected sets outperform fixed baseline choices, including the default $[2, N//2, N-3]$ rule, while providing a clearer explanation of why some defaults can remain competitive. Our results provide a practical recipe for multi-layer hidden-state selection for speculative decoding, insights into the geometry and information in LLM decoder states, and a foundation for extending similar analysis to related applications.

  \end{abstract}

% MAIN PAPER
% ============================================================

% ===========================
% INTRODUCTION
% ===========================
\section{Introduction}
Hidden states in decoder-only LLMs capture the progressive transformation of token representations through depth. While prior studies analyze individual layers through rank, sparsity, and entropy, modern inference pipelines increasingly use \emph{multiple} target-model hidden states jointly. In speculative decoding methods such as EAGLE3, selected target-model layers are fused and provided to the draft model. However, in practice these layers are usually selected by simple fixed rules (e.g., positional heuristics) rather than an explicit objective that balances informativeness and redundancy.

\begin{figure}[ht]
    \centering
    \includegraphics[width=0.95\linewidth]{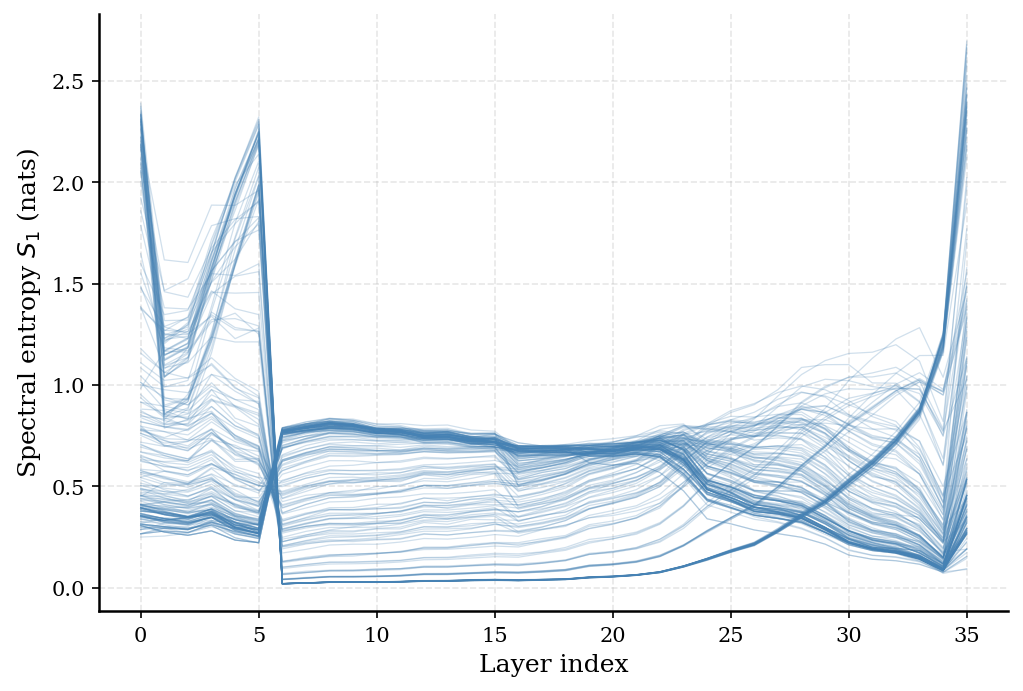}
    \caption{Prompt-wise per-layer Shannon entropy curves for \texttt{Qwen3-4B-Instruct}. The curves show a depth-wise non-uniform profile (an ``activation wave''), with distinct early (0--6), middle (7--28), and late-layer (29--35) regimes.}
    \label{fig:activation-wave-qwen}
\end{figure}

This creates both a practical and scientific gap. Practically, multi-layer selection is combinatorial. For a 36-layer model, choosing only three layers yields thousands of combinations, and the search space grows rapidly for larger tuples. Scientifically, it remains unclear which layer combinations are jointly informative versus geometrically redundant. As a result, default position-based heuristics (uniform selection with the initial and final layers offset) may perform well in some regimes without providing a principled explanation of when or why they work.

In this work, we view multi-layer hidden-state selection through the lenses of information maximization and redundancy minimization. We use matrix entropy to quantify informativeness of layer sets and linear CKA to quantify representational redundancy. We first introduce \textbf{E}ntropy-based \textbf{L}ayer \textbf{S}et \textbf{I}teration (\textbf{ELSI}), a practical entropy-based procedure that scales to realistic model depths. ELSI uses beam search to efficiently identify high-entropy layer sets by evaluating entropy on concatenated hidden states across layers. We then introduce \textbf{ELSI-C}, an entropy-initialized CKA-transition extension that promotes diversity-aware layer expansion. Alongside ELSI/ELSI-C, we introduce token-entropy filtering during calibration: hidden states of low predictive-entropy tokens are removed so that selection statistics focus on significant and informative positions.

We empirically study \texttt{Qwen3-4B-Instruct-2507} and \texttt{Llama-3.1-8B-Instruct}. Our analyses reveal consistent trends across both models: (i) a non-uniform per-layer Shannon entropy profile (an \emph{activation wave} across depth, Fig.~\ref{fig:activation-wave-qwen}), (ii) non-uniform entropy gain under layer concatenation, and (iii) CKA transition boundaries that indicate depth-wise representational compartments. These observations motivate ELSI, an entropy-based selector, and ELSI-C, a diversity-aware extension. Moreover, they clarify why fixed baselines can still be competitive when they overlap with high-scoring entropy-beam regions.

\paragraph{Contributions.}
\begin{itemize}
    \item We propose a principled framework for studying and selecting \emph{multiple} target-model hidden layers using matrix entropy of concatenated representations and CKA transitions between depths. 
    \item We introduce \textbf{ELSI}, a practical entropy-based selector that utilizes beam search to create informative layer sets.
    \item We introduce \textbf{ELSI-C}, an entropy-initialized CKA-transition extension for diversity-aware expansion.
    \item We introduce calibration token-entropy filtering, where hidden states of low predictive-entropy tokens are removed to sharpen layer-discriminative statistics.
    \item We demonstrate practical gains for EAGLE3-style conditioning using ELSI/ELSI-C-selected sets relative to the default $[N-1]$ and $[2, N//2, N-3]$ rules, while improving interpretability.
\end{itemize}

% ===========================
% RELATED WORK
% ===========================
\section{Related Work}
\label{sec:related}
\paragraph{Layer specialization and redundancy in transformers.}
Prior work shows that transformer layers are not uniformly useful and that representational redundancy is often concentrated in middle layers. Transformer layer-function studies report that many middle layers are partially interchangeable, while early and late layers remain more functionally distinct \cite{sun2025transformer}. Related pruning/acceleration work similarly finds that substantial middle-layer computation can be removed with limited quality loss \cite{men2024shortgpt}. Representation-geometry analyses also suggest structured evolution of hidden states across depth \cite{valeriani2023geometry, skean2025layer}. These studies motivate transition-aware reasoning: the depth axis can be divided into coherent compartments separated by sharp representational changes. To algorithmically identify these compartments, we introduce CKA-transition diagnostics.

\paragraph{Spectral and rank-based analyses.}
A complementary line of research studies rank and spectral structure for finetuning and compression. Adaptive rank-allocation methods for LoRA finetuning \cite{hu2022lora} demonstrate that rank requirements vary by layer/module and should not be treated uniformly \cite{zhang2023adalora, kumaravelu2026post}. Other methods use low-rank approximations of transformer components for efficiency \cite{ichi2026low}. Random-matrix and singular-value analyses further show that spectral tails and non-dominant directions can materially affect behavior \cite{staats2026small}. However, these works primarily deal with over-parameterization and compression - they do not address the problem of \emph{selection} of sets of activations.

\paragraph{Speculative decoding and hidden-state conditioning.}
Recent speculative decoding frameworks such as EAGLE \cite{li2024eagle, li2026eagle} and DFlash \cite{chen2026dflash} condition draft models on selected target-model hidden states. In current practice, those target layers are usually chosen by fixed, position-based heuristics (e.g., $[N-1]$ or $[2, N//2, N-3]$) which may lead to redundant representations. Our work addresses this gap through ELSI, which maximizes information and diversity through matrix entropy and CKA-transition structure signals for principled multi-layer selection. We further introduce beam-overlap diagnostics to explain when and why position-based baseline rules still remain competitive.

\paragraph{Positioning of this work.}
% Relative to prior layer-analysis and rank-based compression literature, our contribution is to \emph{jointly} optimize hidden-layer sets for downstream draft conditioning: (i) quantify set informativeness via entropy, (ii) control redundancy via CKA, and (iii) evaluate practical gains in EAGLE3-style speculative decoding with interpretable diagnostics.

Relative to prior layer-analysis and rank-based compression literature, our contribution is to \emph{jointly} optimize hidden-layer sets for downstream draft conditioning: (i) quantify set informativeness via entropy of fused representations with ELSI, (ii) control redundancy via CKA-transition-guided expansion with ELSI-C, and (iii) evaluate practical gains in EAGLE3-style speculative decoding with interpretable diagnostics.

% ===========================
% PRELIMINARIES
% ===========================
\section{Preliminaries}
\label{sec:prelim}

% \subsection{Setup and Notation}
\label{subsec:setup}

Let the target model in the speculative decoding pipeline have $N$ transformer layers indexed by $\{0,\dots,N-1\}$ and hidden dimension $d$. For a sequence of length $T$, let the hidden states output by layer-$l$ be
\begin{equation}
h_l \in \mathbb{R}^{T\times d}.
\label{eq:hidden-layer}
\end{equation}
For any selected layer set $\mathcal{L}\subseteq\{0,\dots,N-1\}$, define the concatenated representation
\begin{equation}
\bar{h}_{\mathcal{L}} = \operatorname{concat}(\{h_l\}_{l\in\mathcal{L}},\text{ feature dim}) \in \mathbb{R}^{T\times |\mathcal{L}|d}.
\label{eq:concat-hidden}
\end{equation}
Following EAGLE3-style conditioning, fused states are down-projected as
\begin{equation}
U_{\mathcal{L}} = \bar{h}_{\mathcal{L}} W_{\text{down}}, \qquad W_{\text{down}}\in\mathbb{R}^{|\mathcal{L}|d\times d}.
\label{eq:downproj}
\end{equation}
Thus, the set of states $\mathcal{L}$ acts as a bottleneck, determining the information seen by the draft model.

\subsection{Matrix Entropy}
\label{subsec:matrix-entropy}

Following \cite{skean2023dime, skean2025layer, giraldo2014measures} for $X\in\mathbb{R}^{T\times D}$, we define $K_X=XX^\top$ and the normalized spectrum
\begin{equation}
p_i(X)=\frac{\lambda_i(K_X)}{\operatorname{tr}(K_X)}=\frac{\sigma_i(X)^2}{\|X\|_F^2},
\label{eq:pi-def}
\end{equation}
where $\sigma_i(X)$ are the singular values. The Rényi-$\alpha$ matrix entropy is
\begin{equation}
S_\alpha(X)=\frac{1}{1-\alpha}\log\sum_i p_i(X)^\alpha, \qquad \alpha>0,\;\alpha\neq1.
\label{eq:renyi-general}
\end{equation}
The limit as $\alpha \to 1$ is 
\begin{equation}
S_1(X)=\lim_{\alpha\to1}S_\alpha(X)=-\sum_i p_i(X)\log p_i(X).
\label{eq:shannon-matrix}
\end{equation}

For a set of hidden states $\mathcal{L}$, we consider the concatenated representation $\bar{h}_{\mathcal{L}}$ defined in Eq.~\eqref{eq:concat-hidden} and compute the matrix entropy as $S_\alpha\left(\bar{h}_{\mathcal{L}}\right)$. Notably, as compared to matrix-based \emph{joint entropy} \cite{skean2023dime}, this formulation of the entropy of a set of hidden states better represents the linear combination of features in the EAGLE-3 style conditioning that we see in Equation~\eqref{eq:downproj}. In our experiments we use $S_1$ to study per-layer trends and select high entropy layers. 

% For current selected set $S$ and candidate layer $k\notin S$, we define marginal gain
% \begin{equation}
% \Delta_{S_2}(S;k)=S_2\!\left(\bar{H}_{S\cup\{k\}}\right)-S_2\!\left(\bar{H}_S\right).
% \label{eq:marginal-entropy-gain}
% \end{equation}
% This leads to pairwise entropy-gain
% \begin{equation}
% \mathrm{EG}_{S_2}(i,j)=S_2([H_; H_j])-S_2(H_i)=\Delta_{S_2}(\{i\};j).
% \label{eq:pairwise-eg}
% \end{equation}
% which we view as heatmaps in our subsequent analysis. 
% Large positive values of entropy-gain imply complementary spectral directions from $j$.

\subsection{Centered-Kernel Alignment (CKA)}
\label{subsec:cka}

For representations $X\in\mathbb{R}^{T\times D_x}$ and $Y\in\mathbb{R}^{T\times D_y}$, we have Gram matrices $ K=XX^\top$ and $L=YY^\top$, with the CKA defined as
\begin{equation}
\operatorname{CKA}(X,Y)=\frac{\langle HKH,HLH\rangle_F}{\|HKH\|_F\,\|HLH\|_F},
\quad H=I-\frac{1}{T}\mathbf{1}\mathbf{1}^\top.
\label{eq:linear-cka}
\end{equation}
% For selected set $L$, average within-set redundancy is
% \begin{equation}
% \overline{\operatorname{CKA}}(L)=\frac{2}{|L|(|L|-1)}\sum_{i<j,\,i,j\in L}\operatorname{CKA}(H_i,H_j).
% \label{eq:avg-cka-set}
% \end{equation}

Let $M_{ij}=\operatorname{CKA}(h_i,h_j)$. We define CKA transition score at boundary between layers $\ell\to\ell+1$ as
\begin{equation}
\delta_\ell=\|M_{\ell+1,:}-M_{\ell,:}\|_2,\qquad \ell=0,\dots,N-2.
\label{eq:cka-transition}
\end{equation}
Large peaks indicate compartment boundaries. The above equation can be extended for concatenated layers as well.

% ===========================
% METHOD
% ===========================
\newcommand{\TODO}[1]{\textcolor{red}{\textbf{TODO: #1}}}
\newcommand{\SA}[1]{\textcolor{blue}{\textbf{SA: #1}}}
\newcommand{\mingu}[1]{\textcolor{purple}{\textbf{mingu: #1}}}
\newcommand{\rg}[1]{\textcolor{magenta}{\textbf{RG: #1}}}
\newcommand{\raghavv}[1]{\textcolor{purple}{\textbf{raghavv: #1}}}
\newcommand{\CITE}{\textcolor{orange}{\textbf{[CITE]}}}
\newcommand{\CITER}[1]{\textcolor{orange}{\textbf{[CITE: #1]}}}

\section{Method}
\label{sec:method}

We propose a three-stage layer-selection pipeline for speculative decoding: (i) token-entropy filtering within hidden states, (ii) hidden-state aggregation to build layer representatives, (iii.a) entropy-driven beam search for informative multi-layer sets (ELSI), and (iii.b) transition-aware CKA expansion for diversity-oriented refinement (ELSI-C). This section formalizes each stage and the evaluation protocol used in our experiments.

\subsection{Token-Entropy Filtering}
\label{subsec:token-filter-method}

For token position $t$, predictive entropy is
\begin{equation}
\mathcal{H}_t = -\sum_{v \in \mathcal{V}} p_t(v)\log p_t(v).
\label{eq:token-entropy}
\end{equation}
We remove low-entropy hidden states within the sequence by top-fraction filtering
\begin{equation}
\Omega = \operatorname{TopFrac}(\{\mathcal{H}_t\}_{t=1}^T,\rho),\quad \rho\in(0,1].
\label{eq:token-topfrac-filter}
\end{equation}
Hidden states restricted to retained positions are denoted $h_l^{(\Omega)}$.
In our experiments, we perform filtering with $\rho=0.3$.

\subsection{Hidden State Collection and Block Averaging}
\label{subsec:collection}

For each model and calibration prompt, we collect hidden states and apply token filtering  ($h_{l}^{(\Omega)}$). We then partition the token axis into non-overlapping blocks of size $B=32$ and average across blocks/samples to obtain layer representatives
\begin{equation}
Z_l\in\mathbb{R}^{B\times d},\qquad l=0,\dots,N-1.
\label{eq:block-avg-rep}
\end{equation}
These $\{Z_l\}$ are used for entropy/CKA analyses and layer-set scoring. In our experiments, we use  200 calibration samples from ShareGPT \cite{sharegpt}.

\subsection{Matrix Entropy-based Expansion (ELSI)}
\label{subsec:entropy-select}

ELSI performs beam search with matrix entropy of concatenated hidden states as the expansion criterion. We use $Z_l$ to denote the layer representatives from Eq.~\eqref{eq:block-avg-rep} and $\bar{Z}_{\mathcal{L}}\triangleq\operatorname{concat}(\{Z_l\}_{l\in \mathcal{L}},\text{ feature dim})$.

ELSI is detailed in Algorithm~\ref{alg:layer-selection-entropy} which begins with entropy-based singleton seeding ([S1]) and then performs beam expansion over set sizes ([S2]). At each depth, candidates are deduplicated ([S3]) and pruned by entropy score ([S4]) before selecting the best size-$m$ set ([S5]).

    % \begin{algorithm}[h]
% \caption{ELSI: Entropy-based Layer Set Iteration}
% \label{alg:layer-selection-entropy}
% \textbf{Input}: Layer-representative hidden states $\{Z_l\}_{l \in \{0,\ldots,N-1\}}$\\
% \textbf{Input}: Number of required layers $k$\\
% \textbf{Output}: Selected layer combination $S$ with $|S|=k$
% \vspace{2pt}
% \hrule
% \vspace{3pt}
% \noindent
% \textit{\# seed with top-$K_{\text{seed}}$ hidden states by entropy} \\
% $\mathcal{S}\gets \arg\text{top-}K_{\text{seed}}\,_{l\in \{0, \cdots, N-1\}} \left(S_1(Z_l)\right)$\\
% \textbf{for} $d \in \{2,\cdots, k\}$ \textbf{do}\\
% \hspace*{1.2em}$\mathit{\mathcal{S}'} \gets [\,]$\\
% \hspace*{1.2em}\textbf{for} $S \in \mathcal{S}$ \textbf{do}\\
% \hspace*{2.4em}\textit{\# expand each candidate with top entropy-scoring additions} \\
% \hspace*{2.4em} $\mathcal{S}' \gets \mathcal{S}' \; \cup \; \arg\text{top-}K_{\text{expand}}\,_{S\cup\{l\}} \left(S_1(\bar{Z}_{S\cup\{l\}})\right)$ \\
% \hspace*{1.2em}\textbf{end for}\\
% \hspace*{1.2em}\textit{\# prune to top-$K_{\text{beam}}$ candidates per level} \\
% \hspace*{1.2em} $\mathcal{S} \gets \arg\text{top-}K_{\text{beam}}\,_{S\in \mathcal{S'}} \left(S_1(\bar{Z}_S)\right)$ \\
% \textbf{end for}\\
% \textit{\# return the top candidate of the final level} \\
% $S^* = \arg \text{top-1}_{S\in \mathcal{S}}(S_1(\bar{Z}_S))$\\
% \textbf{return} $S^*$
% \end{algorithm}

\begin{algorithm}[h]
\caption{ELSI: Entropy-based Layer Set Iteration}
\label{alg:layer-selection-entropy}
\textbf{Input}: Layer representatives $\{Z_l\}_{l=0}^{N-1}$, target set size $m$\\
\textbf{Input}: Beam hyperparameters $( K_{\text{expand}}, K_{\text{beam}})$\\
\textbf{Output}: Selected layer set $\mathcal{L}^\star$ with $|\mathcal{L}^\star|=m$
\vspace{2pt}
\hrule
\vspace{3pt}
\noindent
\textit{[S1] initial beam: top-entropy singletons}\\
$\mathcal{B}_1\gets \arg\text{top-}K_{\text{beam}}\,_{l\in\{0,\ldots,N-1\}}\; S_1(Z_l)$\\
\textit{[S2] entropy-driven expansion across depths}\\
\textbf{for} $d\in\{2,\ldots,m\}$ \textbf{do}\\
\textit{\# $d=2$ forms pairs; $d=3$ forms triples, etc.}\\
\hspace*{1.2em}$\mathcal{C}_d\gets[\,]$\\
\hspace*{1.2em}\textbf{for each} $\mathcal{L}\in\mathcal{B}_{d-1}$ \textbf{do}\\
\hspace*{2.4em}\textit{\# top extension candidates}\\
\hspace*{2.4em}$\mathcal{E}(\mathcal{L})\gets \arg\text{top-}K_{\text{expand}}\,_{l\notin \mathcal{L}}\; S_1(\bar{Z}_{\mathcal{L}\cup\{l\}})$\\
\hspace*{2.4em}$\mathcal{C}_d\gets \mathcal{C}_d\cup\{\mathcal{L}\cup\{l\}: l\in\mathcal{E}(\mathcal{L})\}$\\
\hspace*{1.2em}\textbf{end for}\\
\hspace*{1.2em}\textit{[S3] deduplicate identical sets from different parents}\\
\hspace*{1.2em}$\mathcal{C}_d\gets\operatorname{Dedup}(\mathcal{C}_d)$\\
\hspace*{1.2em}\textit{[S4] beam prune by set entropy}\\
\hspace*{1.2em}$\mathcal{B}_d\gets \arg\text{top-}K_{\text{beam}}\,_{\mathcal{L}\in\mathcal{C}_d}\; S_1(\bar{Z}_{\mathcal{L}})$\\
\textbf{end for}\\
\textit{[S5] final selection at target size $m$}\\
$\mathcal{L}^\star\gets \arg\max_{\mathcal{L}\in\mathcal{B}_m} S_1(\bar{Z}_{\mathcal{L}})$\\
\textbf{return} $\mathcal{L}^\star$
\end{algorithm}

A visualization of the beams produced by ELSI using $K_{\text{beam}}=10$ is shown in Supplementary Fig. S3.
 
\subsection{ELSI-C: Entropy-Initialized CKA Expansion}
\label{subsec:hybrid-select}

ELSI-C reuses entropy initialization and entropy scoring, but restricts each expansion to a transition-aware candidate pool from conditional CKA.  Algorithm~\ref{alg:layer-selection-elsic} first initializes the beam using entropy ([C1]). For each beam member, it computes a transition score for each candidate using CKA ([C2]). Candidate layers with the highest transition scores are then kept for each beam ([C3]). Next, candidate expansions are aggregated with deduplication and the beam is pruned based on entropy ([C4]). The final size-$m$ set is selected as the top beam element ([C5]).

\begin{algorithm}[h]
\caption{ELSI-C: Entropy-based Layer Set Iteration with CKA Filtering}
\label{alg:layer-selection-elsic}
\textbf{Input}: Layer representatives $\{Z_l\}_{l=0}^{N-1}$, target set size $m$\\
\textbf{Input}: Hyperparameters $(K_{\text{beam}},K_{\text{pool}})$\\
\textbf{Output}: Selected set $\mathcal{L}^\star$ with $|\mathcal{L}^\star|=m$
\vspace{2pt}
\hrule
\vspace{3pt}
\noindent
\textit{[C1] initial beam: top-entropy singletons}\\
$\mathcal{B}_1\gets \arg\text{top-}K_{\text{beam}}\,_{l}\;S_1(Z_l)$\\
\textbf{for} $d\in\{2,\ldots,m\}$ \textbf{do}\\
\textit{\# $d=2$ forms pairs; $d=3$ forms triples, etc.}\\
\hspace*{1.2em}$\mathcal{C}_d\gets[\,]$\\
\hspace*{1.2em}\textbf{for each} $\mathcal{L}\in\mathcal{B}_{d-1}$ \textbf{do}\\
\hspace*{2.4em}\textit{\# conditional CKA matrix}\\
\hspace*{2.4em}$M_{\mathcal{L}}(i,j)\gets\operatorname{CKA}(\bar{Z}_{\mathcal{L}\cup\{i\}},\bar{Z}_{\mathcal{L}\cup\{j\}})$\\
\hspace*{2.4em}\textit{[C2] calculate transition scores for each candidate}\\
\hspace*{2.4em}$\delta_{\mathcal{L}}(r)\gets\|M_{\mathcal{L}}(r,:)-M_{\mathcal{L}}(r-1,:)\|_2$\\
% \hspace*{2.4em}$\tau\gets\eta\max_r\delta_{\mathcal{L}}(r)$; pick up to $n_{\text{select}}$ peaks above $\tau$ with gap $g$\\
\hspace*{2.4em}\textit{[C3] candidate layers with highest transition scores}\\
% \hspace*{2.4em}$\mathcal{P}_{\mathcal{L}}\gets$ selected transition layers not in $\mathcal{L}$\\
\hspace*{2.4em}$\mathcal{Q}_{\mathcal{L}}\gets\arg\text{top-}K_{\text{pool}}\,_{l\notin\mathcal{L}}\;\delta_{\mathcal{L}}(l)$\\
\hspace*{2.4em}$\mathcal{C}_d\gets \mathcal{C}_d\cup\{\mathcal{L}\cup\{l\}: l\in\mathcal{Q}_{\mathcal{L}}\}$\\
% \hspace*{2.4em}\textit{[C5] entropy scoring within filtered pool}\\
% \hspace*{2.4em}\textbf{for each} $l\in\mathcal{Q}_{\mathcal{L}}$ \textbf{do} add candidate $(\mathcal{L}\cup\{l\},\;S_1(\bar{Z}_{\mathcal{L}\cup\{l\}}))$ to $\mathcal{C}_d$ \textbf{end for}\\
\hspace*{1.2em}\textbf{end for}\\
\hspace*{1.2em}\textit{[C4] deduplicate + beam prune with entropy}\\
\hspace*{1.2em}$\mathcal{C}_d\gets\operatorname{Dedup}(\mathcal{C}_d)$\\
\hspace*{1.2em}$\mathcal{B}_d\gets\arg\text{top-}K_{\text{beam}}\,_{\mathcal{L}\in\mathcal{C}_d}\;S_1(\bar{Z}_{\mathcal{L}})$\\
\textbf{end for}\\
\textit{[C5] final selection at target size $m$}\\
$\mathcal{L}^\star\gets \arg\max_{\mathcal{L}\in\mathcal{B}_m} S_1(\bar{Z}_{\mathcal{L}})$\\
\textbf{return} $\mathcal{L}^\star$
\end{algorithm}

% ===========================
% RESULTS
% ===========================
\section{Experiments}
\label{sec:results}
We use open-source models from two different families: \texttt{Qwen3-4B-Instruct-2507} \cite{yang2025qwen3} and \texttt{Llama3.1-8B-Instruct} \cite{grattafiori2024llama} to analyze the per-layer entropy (\cref{subsec:results-shannon}), concatenated entropy (\cref{subsec:results-eg}), and CKA transitions across layers (\cref{subsec:results-cka}). We additionally include the layer candidates selected using ELSI and ELSI-C for these models (\cref{subsec:results-selection}).   

\subsection{Per-layer matrix entropy}
\label{subsec:results-shannon}

\begin{figure}[ht]
    \centering
    \includegraphics[width=0.95\linewidth]{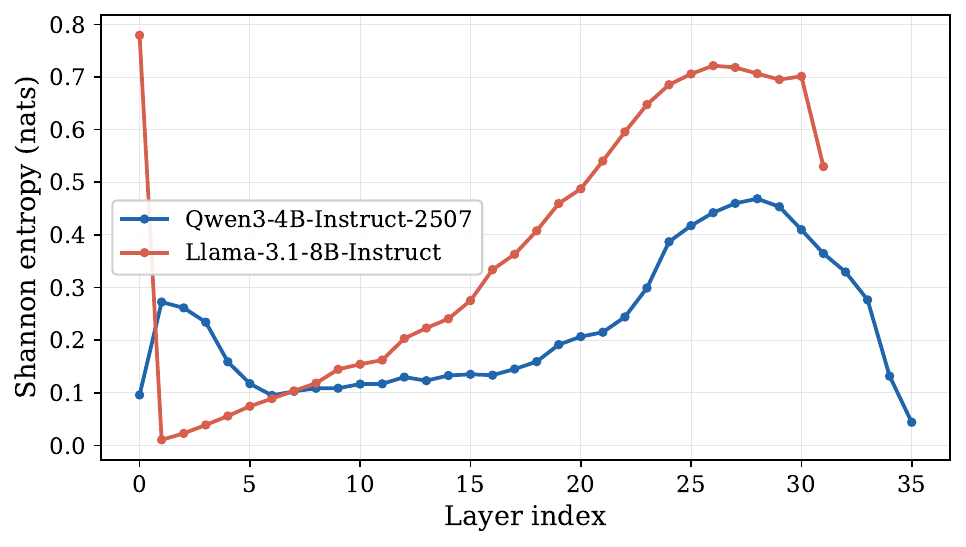}
    \caption{Per-layer Shannon entropy ($S_1$) across depth for Qwen3-4B-Instruct-2507 and Llama-3.1-8B-Instruct.}
    \label{fig:entropy-shannon}
\end{figure}

We first analyze per-layer Shannon matrix entropy $S_1$. In Figure~\ref{fig:entropy-shannon}, we see that across models, entropy varies substantially with depth, indicating non-uniform layer informativeness. \texttt{Llama3.1-8B-IT} shows high entropy for the first layer which then drops sharply and then rises, peaking around layer $27$. Based on this we add another condition when selecting a single layer: in addition to having high-entropy, the layer index should occur after the middle layer index. \texttt{Qwen3-4B-IT} shows high-low-high entropy peaking at layer $28$. While per-layer entropy can inform the selection of a single layer, it does not immediately reveal favorable \emph{sets} of layers, motivating deeper analysis. 

\subsection{Entropy of layer pairs}
\label{subsec:results-eg}
\begin{figure*}[t]
    \centering
    \begin{subfigure}[b]{0.48\linewidth}
        \includegraphics[width=\linewidth]{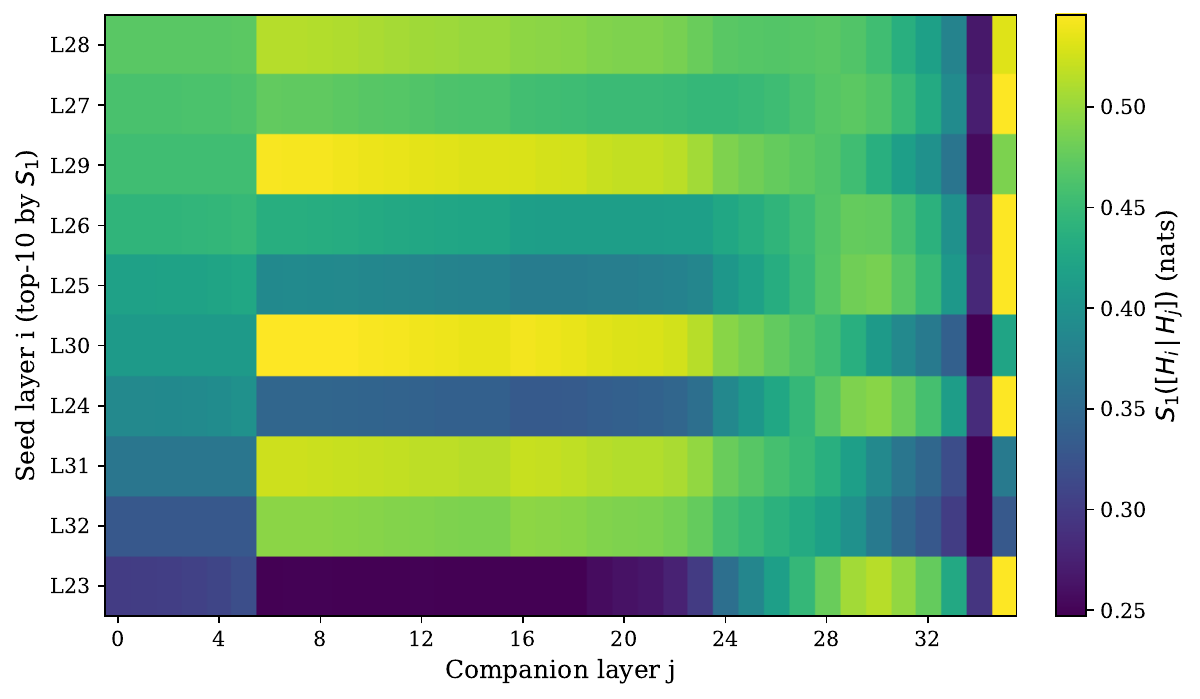}
        \caption{Qwen3-4B-Instruct-2507}
    \end{subfigure}
    \hfill
    \begin{subfigure}[b]{0.48\linewidth}
        \includegraphics[width=\linewidth]{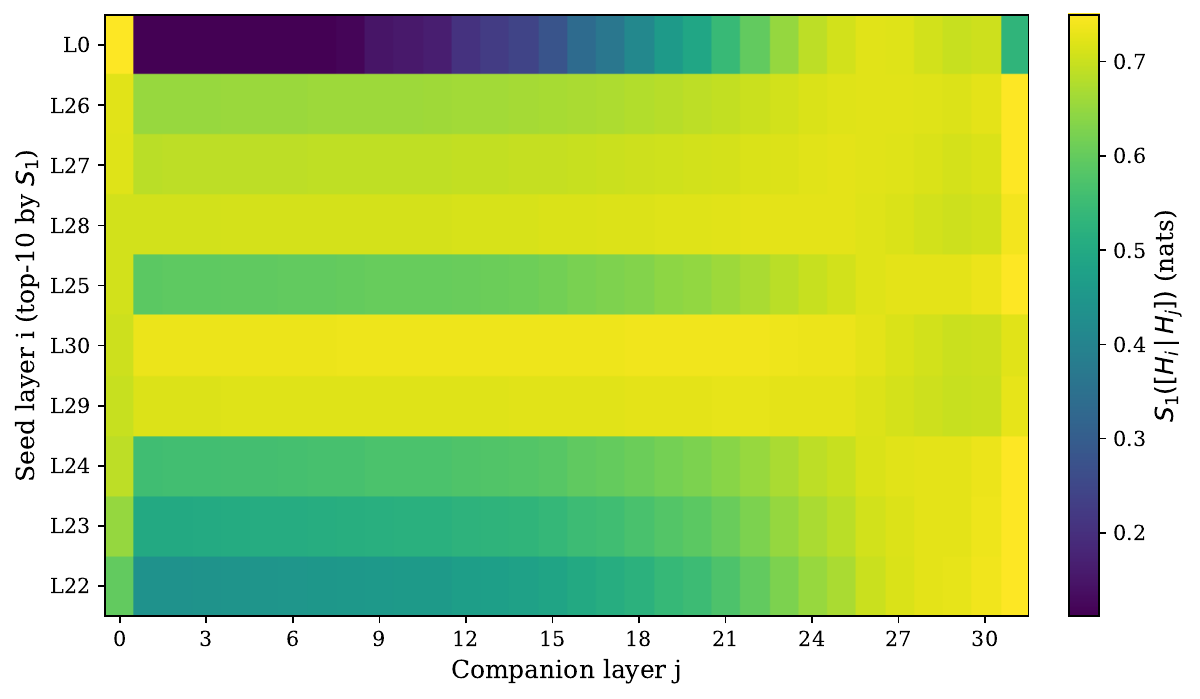}
        \caption{Llama-3.1-8B-Instruct}
    \end{subfigure}
    \caption{Entropy of layer pairs. Top-$k$ layers by entropy are chosen as seeds and paired with the remaining layers $S_1(\{H_i,H_j\})$.}
    \label{fig:eg-topk-joint}
\end{figure*}

We next evaluate the entropy of concatenations of layer pairs. Figure~\ref{fig:eg-topk-joint} visualizes top-$k$ seed layers (by per-layer entropy) against all companion layer concatenations. We see non-uniform gain patterns: some layer pairs provide strong complementary directions while others are largely redundant. Moreover, different seed layers benefit from different companion layers. These results justify the need to optimize sets of layers jointly and entropy of concatenated representations as a metric to achieve this. 
% Both models show different trends, with \texttt{Qwen3-4B-IT} showing high entropy for few layers with decreasing entropy towards later layers, \texttt{Llama3.1-8B-IT} show higher entropy as we go deeper. 
\subsection{Token-Entropy Filtering}
\label{subsec:token-filter}
\begin{figure}[h]
    \centering
        \includegraphics[width=0.9\linewidth]{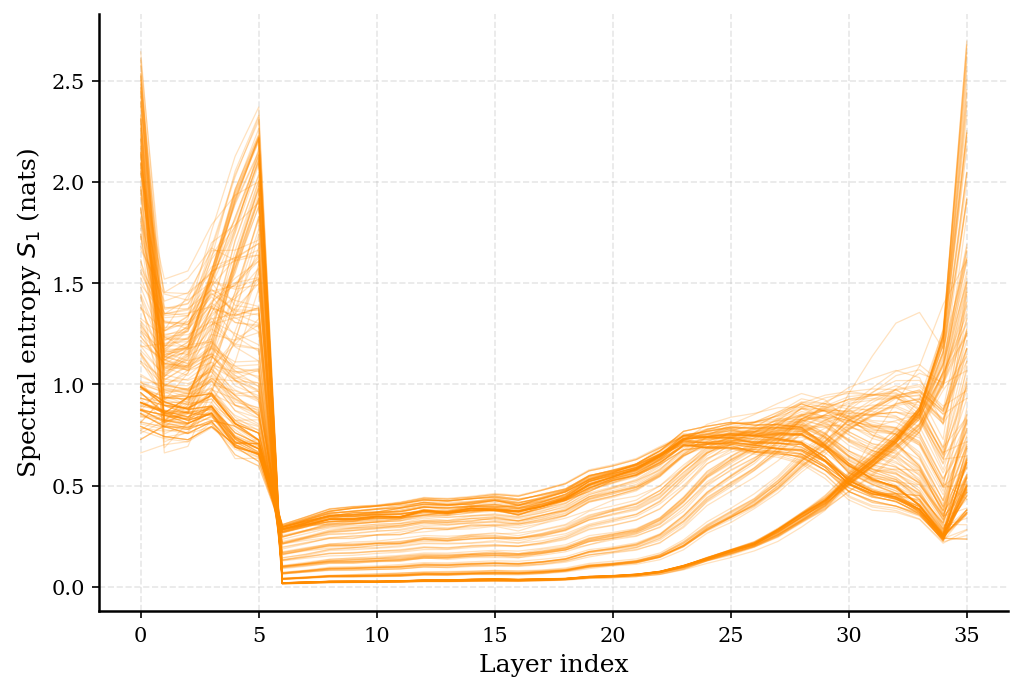}
        \caption{Token-entropy filtered prompt-token set ($\rho=0.30$) lowers the entropy variance across prompts while preserving the activation-wave shape.}
    \label{fig:activation-wave-default-vs-filter}
\end{figure}

Across models, entropy varies substantially with depth and also across prompts as seen in Fig. \ref{fig:activation-wave-qwen} where the entropy variance across prompts is very high for initial and final layers. This can result in a prompt dependent layer ordering which is infeasible as the set of layers selected remains fixed through training and inference. 
We therefore use token filtering which reduces the cross-prompt entropy variance by $48.4\%$ (early layers $0$--$6$), $45.8\%$ (middle layers $7$--$28$), and $35.6\%$ (late layers $29$--$35$), while preserving the same depth-wise activation-wave profile (Fig.~\ref{fig:activation-wave-default-vs-filter}).

% \begin{figure*}[t]
%     \centering
%     \begin{subfigure}[b]{0.48\linewidth}
%         \includegraphics[width=\linewidth]{figures-v2/pairwise_eg_s2_qwen3-4b.pdf}
%         \caption{Qwen3-4B-Instruct-2507}
%     \end{subfigure}
%     \hfill
%     \begin{subfigure}[b]{0.48\linewidth}
%         \includegraphics[width=\linewidth]{figures-v2/pairwise_eg_s2_llama-3.1-8b.pdf}
%         \caption{Llama-3.1-8B-Instruct}
%     \end{subfigure}
%     \caption{Pairwise Rényi-2 entropy gain heatmaps.}
%     \label{fig:eg-pairwise}
% \end{figure*}

% \begin{figure*}[t]
%     \centering
%     \begin{subfigure}[b]{0.48\linewidth}
%         \includegraphics[width=\linewidth]{figures-v2/topk_joint_entropy_s2_qwen3-4b_token_filter_ff30.pdf}
%         \caption{Qwen3-4B-Instruct-2507}
%     \end{subfigure}
%     \hfill
%     \begin{subfigure}[b]{0.48\linewidth}
%         \includegraphics[width=\linewidth]{figures-v2/topk_joint_entropy_s2_llama-3.1-8b_token_filter_ff30.pdf}
%         \caption{Llama-3.1-8B-Instruct}
%     \end{subfigure}
%     \caption{Entropy of layer pairs. Top-$k$ layers by entropy are chosen as seeds and paired with the remaining layers $S_2([H_i;H_j])$.}
%     \label{fig:eg-topk-joint}
% \end{figure*}

\subsection{CKA compartments and transitions}
\label{subsec:results-cka}

\begin{figure}[!htbp]
    \centering
    \begin{subfigure}[b]{0.90\linewidth}
        \includegraphics[width=\linewidth]{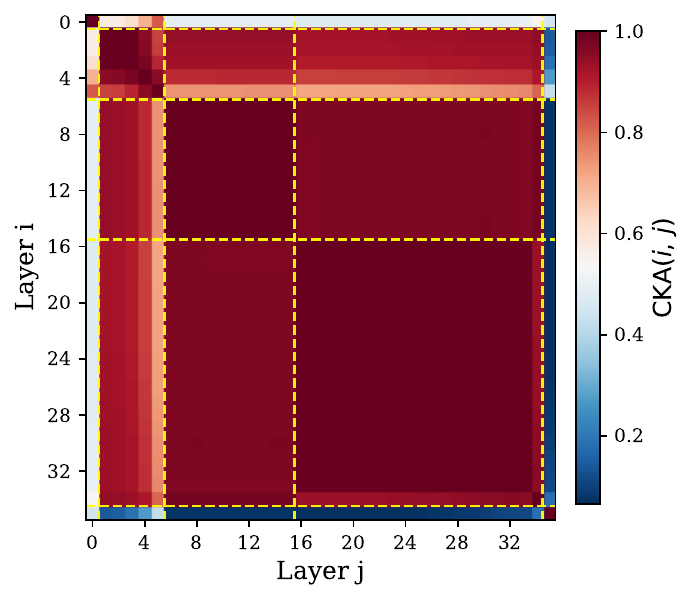}
        \caption{Qwen3-4B-Instruct-2507}
    \end{subfigure}
    \begin{subfigure}[b]{0.90\linewidth}
    \includegraphics[width=\linewidth]{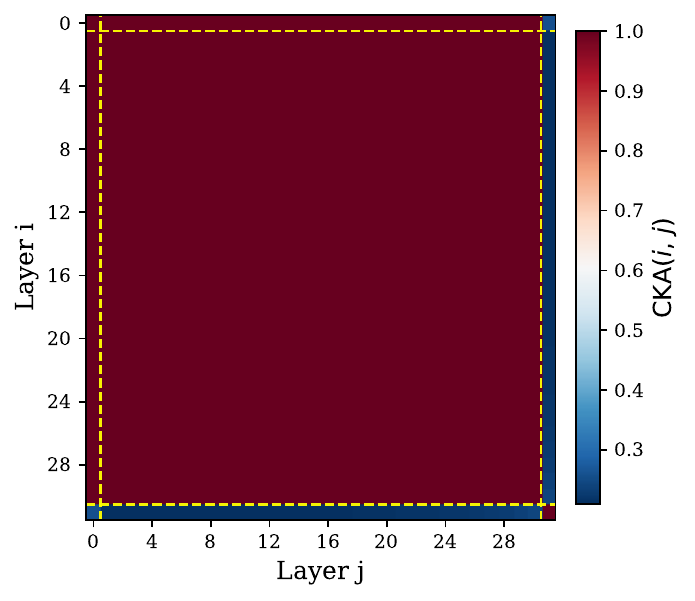}
        \caption{Llama-3.1-8B-Instruct}
    \end{subfigure}
    \caption{Pairwise CKA with calculated transition boundary overlaid in yellow.}
    \label{fig:cka-pairwise}
\end{figure}

Figure~\ref{fig:cka-pairwise} shows pairwise CKA with transition boundaries derived from Eq.~\ref{eq:cka-transition} overlaid. 
The transition analysis reveals clear depth-wise representational compartments. \texttt{Qwen3-4B-IT} has more transitions than \texttt{Llama3.1-8B-IT}, showing more variation as we go deeper into the network, with two large compartments in the middle. \texttt{Llama3.1-8B-IT} on the other hand has similar CKA throughout except for the first and last layers. 

From this perspective, EAGLE2-like methods which only used last layer hidden-states might miss out on other high information layers. EAGLE3 which selects layers following $[2, N//2, N-3]$ selects layers in different compartments, thus resulting in improved performance over EAGLE2, though it might still be sub-optimal by having overlapping information.

% \begin{figure}[H]
%     \centering
%     \begin{subfigure}[b]{0.80\linewidth}
%         \includegraphics[width=\linewidth]{figures-v2/cka_transition_qwen3-4b_token_filter_ff30_msf001.pdf}
%         \caption{Qwen3-4B-Instruct-2507}
%     \end{subfigure}
%     \begin{subfigure}[b]{0.80\linewidth}
%     \includegraphics[width=\linewidth]{figures-v2/cka_transition_llama-3.1-8b_token_filter_ff30_msf001.pdf}
%         \caption{Llama-3.1-8B-Instruct}
%     \end{subfigure}
%     \caption{Pairwise CKA with calculated transition boundary overlaid in yellow.}
%     \label{fig:cka-pairwise}
% \end{figure}

\subsection{Selection tables and baseline intuition}
\label{subsec:results-selection}
We show in Table~\ref{tab:entropy-no-mcol} the layers selected by ELSI and ELSI-C when choosing singletons, pairs, and triples of layers. We also show how the entropy of the layer sets selected by ELSI compares to the entropy of the layer choices of EAGLE2 and EAGLE3. We present this for both \texttt{Qwen3-4B-IT} and \texttt{Llama3.1-8B-IT} models and observe that layers selected using ELSI and ELSI-C always have higher entropy than those selected naively in EAGLE2 and EAGLE3. 
% A detailed table with the global entropy-based ranks of each layer set is shown in Supplementary Table S1.
\begin{table}[t]
\centering
\small
\setlength{\tabcolsep}{0.9mm}
% \renewcommand{\arraystretch}{1.10}
% \resizebox{\columnwidth}{!}{%
\begin{tabular}{l c c c c}
\toprule
\textbf{Method} &
\multicolumn{2}{c}{\textbf{Qwen3-4B-Instruct}} &
\multicolumn{2}{c}{\textbf{Llama-3.1-8B-Instruct}} \\
\cmidrule(lr){2-3}\cmidrule(lr){4-5}
& \textbf{Layers selected} & \textbf{Entropy} & \textbf{Layers selected} & \textbf{Entropy} \\
\midrule
\multicolumn{5}{l}{\textbf{m=1}} \\
EAGLE2  & \{L35\} & 0.044 & \{L31\} & 0.530 \\
ELSI    & \{L28\} & 0.469 & \{L26\}  & 0.718 \\
ELSI-C  & \{L28\} & 0.469 & \{L26\}  & 0.718 \\
\midrule
\multicolumn{5}{l}{\textbf{m=2}} \\
ELSI    & \{L23, L35\} & 0.625 & \{L22, L31\} & 0.824 \\
ELSI-C  & \{L25, L35\} & 0.588 & \{L25, L31\} & 0.797 \\
\midrule
\multicolumn{5}{l}{\textbf{m=3}} \\
EAGLE3  & \{L2, L18, L33\} & 0.441 & \{L2, L16, L29\} & 0.687 \\
ELSI    & \{L5, L23, L35\} & 0.627 & \{L1, L22, L31\} & 0.910 \\
ELSI-C  & \{L6, L28, L35\} & 0.593 & \{L1, L25, L31\} & 0.889 \\
\bottomrule
\end{tabular}%
% }

\caption{Layer selections and entropy for Qwen3-4B-Instruct and Llama-3.1-8B-Instruct, grouped by selection size $m$.}
\label{tab:entropy-no-mcol}
\end{table}

\subsection{Impact on Speculative Decoding}
\label{subsec:results-specdec}

\subsubsection{Baselines and Evaluation.}
\label{subsubsec:baselines}
We train EAGLE-style draft models using ELSI and ELSI-C and compare against layer selections based on: (i) EAGLE2 $[N-1]$, (ii) EAGLE3 $[2,N//2,N-3]$, and (iii) low-entropy negative sets.

\subsubsection{Training and Evaluation Setup.}
\label{subsubsec:training-details}
All the draft models were trained using a distillation dataset generated using ShareGPT \cite{sharegpt} and Ultrachat \cite{ding2023enhancing} on 8 H100 GPUs. For evaluation we ablate over (a) number of draft tree nodes: 30 and 60, and (b) temperature sampling: T=0.0 (greedy) and T=1.0. We evaluate on SpecBench \cite{xia-etal-2024-unlocking}, which covers a wide range of tasks and report average accepted tokens ($\tau$). Note that average accepted tokens will directly lead to higher token per second in this scenario for same number of hidden-states chosen. For a smaller set of target states, we expect higher throughput due to the usage of a smaller down-projection layer of size $|\mathcal{L}|d \times d$, though this might not be visible on GPUs which are not memory-bound. We leave it to future work to experiment with ELSI-generated layer sets on edge-devices. 

We show our main results for \texttt{Qwen3-4B-IT} in Table~\ref{tab:qwen3-4b_block_efficiency_combined} and make several observations in this data:
\begin{itemize}
    \item ELSI ($m{=}3$), matching the number of EAGLE3 layers, is \textbf{the
    best performing method on average across all settings}, improving
    over EAGLE3 by $1.6$--$2.2\%$ ($+0.085\,\tau$ on average) and over
    EAGLE2 by $6.1$--$7.3\%$.
    
    \item Across all four settings, ELSI ($m{=}1$) matches or exceeds EAGLE3 on average ($+0.01$ to $+0.04$ $\tau$), showing that \textbf{a single well-chosen layer
    matches EAGLE3's default 3-layer heuristic}.

    \item \textbf{ELSI tends to provide the most benefit in categories where
    EAGLE3 performs weakest}, for example Translation ($2.6$--$5.6\%$)
    and Summarization ($2.1$--$4.2\%$), but can also improve acceptance
    rate significantly in performant categories such as STEM
    ($2.4$--$3.9\%$).

    \item \textbf{ELSI provides consistent benefits across sampling temperatures and draft tree sizes configurations}, and thus may be applied to diverse deployment scenarios.
\end{itemize}

% Impressively, using a \emph{single layer} selected using ELSI outperforms both EAGLE2 and EAGLE3 across draft nodes and temperature settings. So ELSI(m=1) beats EAGLE2 consistently by ${\sim}5\%$, beats the low-entropy negative control by ${\sim}4{-}4.8\%$, and EAGLE3 by ${\sim}0.4{-}0.9\%$ — \textbf{a single well-chosen layer matches EAGLE3's default 3-layer heuristic}.
% Expanding to m=2 roughly doubles-to-triples ELSI's edge over EAGLE3 (to ${\sim}1.4-2.1\%$), and the gain is largest under stochastic sampling (T=1). With m=3, now matching the number of hidden layers used by EAGLE3, ELSI emerges as the best performing method on average. From Table~\ref{tab:entropy-no-mcol}, we observe that ELSI selects different first and second layer compared to EAGLE3 with third layer being in close proximity. 

We also include a low-entropy control experiment wherein we choose the layer set with the lowest entropy resulting in layers $\{0, 1, 35\}$). We train an EAGLE-3 model with this configuration and see that it consistently results in the lowest acceptance rate. This shows that using low entropy hidden-states (which could have overlapping information) doesn't provide a strong conditioning to the draft model. 

% We make several observations in this data: (i) Across settings, ELSI (m=1) outperforms EAGLE3 on average, showing that \textbf{a single well-chosen layer matches EAGLE3's default 3-layer heuristic}. (ii) ELSI (m=3), matching the number of EAGLE3 layers, is the best performing method on average across all settings (iii) ELSI's tends to provide the most benefit in categories where EAGLE3 performs weakest, for example Writing and Translation, but can also improve acceptance rate significantly in performant categories such as STEM. (iv) ELSI provides consistent benefits across sampling temperatures and draft tree sizes and thus may be applied to diverse deployment scenarios.

% \SA{this needs revision it only talks about m=1}
% We also make several category-wise observations as follows: (i) For Translation, ELSI (m=1) beats EAGLE3 by $0.19 {-} 0.25$ $\tau$ (+$5 {-} 7\%$) in every configuration, and beats EAGLE2 by $0.45 {-} 0.62$ $\tau$. (ii) Extraction shows large gains of ELSI (m=1) compared to EAGLE2 ($+0.62 \, \tau$ at $\{30, T=1.0\}$) and the low-entropy control (+$0.47 \, \tau$ at $\{60, T=1.0\}$) but slight worst performance compared to EAGLE3, 
%(iii) \textbf{Coding} seems to be an outlier where EAGLE3 (60, T=1.0) performs better than ELSI(m=3) by a margin.  

Table~\ref{tab:block_efficiency_llama_combined} reports the acceptance rate for \texttt{Llama3.1-8B-IT}. We show a subset of SpecBench
categories (CODING, QA, ROLEPLAY, WRITING, M/R), and report the average (AVG) across all categories. Entropy-guided selection remains highly efficient on this model: \textbf{ELSI ($m{=}1$) exceeds EAGLE2 by $2.8$--$4.9\%$ while attaining ${\sim}99.0\%$ of EAGLE3's block efficiency using a single hidden state rather than three.} However, increasing the number of selected layers yields no further gain, with AVG varying by only $0.017\,\tau$ across $m=1,2,3$. On average, ELSI ($m{=}3$) trails EAGLE3 by $1.0\%$. However, at $T{=}1.0$, ELSI ($m{=}3$) improves over EAGLE3 on Roleplay by $4.7$--$5.9\%$ and on Coding by $3.0$--$3.3\%$, whereas at $T{=}0.0$ the two are closer. The corresponding losses on Writing at $T{=}1.0$ leave the average unchanged.

We attribute the reduced headroom overall to the greater interchangeability of Llama's decoder layers, visible as broad regions of high pairwise CKA in Figure~\ref{fig:cka-pairwise}. When neighboring layers encode similar information, the choice of layers matters correspondingly less, and a uniformly spaced selection already performs well. Framing layer selection through ELSI thus allows us to identify these differing regimes across models: where
representations are diverse, selection results in significant benefits, and where they are redundant, it instead allows for a smaller layer-set selection. 

% Table~\ref{tab:block_efficiency_llama_combined} reports average accepted length across draft-tree size and temperature for \texttt{Llama3.1-8B-IT}. Due to page limits, we show a subset of categories (CODING, QA, ROLEPLAY, WRITING, M/R, AVG).
% Unlike \texttt{Qwen}, where ELSI/ELSI-C modestly outperform EAGLE3, \texttt{Llama3.1-8B-IT} shows EAGLE3 as the strongest method by AVG in every (Nodes, $T$) block; however, ELSI still improves over EAGLE2 by 2.6--5.0\% AVG throughout. The
% EAGLE3--ELSI gap on M/R narrows substantially at $T{=}1$ and reverses at Nodes$=60$, indicating that ELSI’s math/reasoning disadvantage is temperature-dependent. ROLEPLAY and WRITING are comparatively volatile at $T{=}1$, with no method
% consistently dominating ROLEPLAY at Nodes$=60$. A plausible explanation is that the hidden-state space of \texttt{Llama3.1-8B-IT} is less compartmentalized across depth, so EAGLE3’s heuristic selection already spans sufficiently diverse states.
% This interpretation is consistent with Table~\ref{tab:entropy-no-mcol}: EAGLE3 and ELSI select nearby early/late layers and differ mainly in the middle layer.

\begin{table*}[!htbp]
\centering
\setlength{\tabcolsep}{0.95mm}
\small
% \resizebox{\textwidth}{!}{
\begin{tabular}{lcccccccccccccc}
\toprule
\textbf{Method} & \textbf{Nodes} & \textbf{T} & CODING & EXTR. & HUMAN. & QA & RAG & ROLE. & STEM & SUMM. & TRANSL. & WRITING & M/R & \textbf{AVG} \\
\midrule
EAGLE2 & \multirow{8}{*}{30} & \multirow{8}{*}{0.0} & 4.73 & 4.21 & 3.69 & 3.90 & 3.94 & 3.67 & 3.98 & 3.38 & 3.39 & 3.95 & 4.91 & 4.00 \\
ELSI (m=1) &  &  & 4.83 & 4.79 & 3.80 & 3.97 & 4.17 & 3.63 & 4.14 & 3.56 & \textbf{3.85} & 4.06 & 5.21 & 4.24 \\
\cmidrule{4-15}
ELSI (m=2) &  &  & 4.90 & 4.61 & 3.86 & 4.05 & \textbf{4.20} & 3.75 & 4.22 & 3.58 & 3.83 & 4.17 & \textbf{5.26} & 4.28 \\
ELSI-C (m=2) &  &  & \textbf{4.93} & 4.77 & 3.82 & 4.07 & 4.18 & 3.76 & 4.22 & 3.59 & 3.76 & 4.19 & 5.20 & 4.26 \\
\cmidrule{4-15}
Low-Entropy &  &  & 4.74 & 4.41 & 3.71 & 3.90 & 4.03 & 3.63 & 4.07 & 3.38 & 3.42 & 4.00 & 4.96 & 4.04 \\
EAGLE3 &  &  & 4.88 & 4.75 & 3.79 & 4.02 & 4.13 & 3.73 & 4.14 & 3.52 & 3.60 & 4.07 & 5.18 & 4.20 \\
ELSI (m=3) &  &  & 4.90 & \textbf{4.80} & \textbf{3.87} & \textbf{4.08} & 4.20 & \textbf{3.86} & \textbf{4.30} & \textbf{3.61} & 3.80 & \textbf{4.22} & 5.25 & \textbf{4.29} \\
ELSI-C (m=3) &  &  & 4.92 & 4.70 & 3.85 & 4.05 & 4.15 & 3.76 & 4.19 & 3.56 & 3.71 & 4.16 & 5.22 & 4.24 \\
\midrule
EAGLE2 & \multirow{8}{*}{30} & \multirow{8}{*}{1.0} & 4.51 & 4.41 & 3.37 & 3.59 & 3.76 & 3.44 & 3.91 & 3.26 & 3.35 & 3.52 & 4.85 & 3.86 \\
ELSI (m=1) &  &  & 4.55 & \textbf{5.03} & 3.48 & 3.60 & 3.98 & 3.31 & 3.95 & 3.41 & \textbf{3.81} & 3.63 & 5.06 & 4.06 \\
\cmidrule{4-15}
ELSI (m=2) &  &  & \textbf{4.68} & 4.88 & \textbf{3.51} & 3.72 & \textbf{4.02} & 3.44 & 4.04 & 3.45 & 3.79 & 3.75 & 5.13 & 4.11 \\
ELSI-C (m=2) &  &  & 4.68 & 4.82 & 3.49 & 3.74 & 3.97 & 3.42 & 3.94 & \textbf{3.47} & 3.74 & 3.70 & 5.13 & 4.09 \\
\cmidrule{4-15}
Low-Entropy &  &  & 4.50 & 4.56 & 3.39 & 3.66 & 3.86 & 3.31 & 3.87 & 3.26 & 3.37 & 3.62 & 4.84 & 3.89 \\
EAGLE3 &  &  & 4.52 & 4.81 & 3.42 & 3.65 & 3.96 & 3.37 & 3.93 & 3.38 & 3.59 & 3.62 & 5.10 & 4.02 \\
ELSI (m=3) &  &  & 4.67 & 4.96 & 3.50 & \textbf{3.74} & 4.01 & 3.34 & \textbf{4.06} & 3.45 & 3.74 & \textbf{3.91} & \textbf{5.17} & \textbf{4.11} \\
ELSI-C (m=3) &  &  & 4.48 & 4.85 & 3.50 & 3.74 & 4.00 & \textbf{3.50} & 3.89 & 3.44 & 3.71 & 3.70 & 5.11 & 4.08 \\
\midrule
EAGLE2 & \multirow{8}{*}{60} & \multirow{8}{*}{0.0} & 5.03 & 4.54 & 4.04 & 4.19 & 4.22 & 3.89 & 4.36 & 3.61 & 3.57 & 4.27 & 5.23 & 4.28 \\
ELSI (m=1) &  &  & 5.17 & \textbf{4.97} & 4.07 & 4.26 & 4.45 & 4.01 & 4.40 & 3.81 & 4.02 & 4.37 & 5.49 & 4.50 \\
\cmidrule{4-15}
ELSI (m=2) &  &  & 5.23 & 4.85 & 4.17 & 4.37 & 4.48 & \textbf{4.21} & 4.49 & 3.82 & 3.97 & \textbf{4.52} & \textbf{5.56} & 4.55 \\
ELSI-C (m=2) &  &  & \textbf{5.23} & 4.94 & \textbf{4.18} & 4.37 & 4.43 & 4.04 & 4.55 & 3.83 & 3.96 & 4.48 & 5.54 & 4.54 \\
\cmidrule{4-15}
Low-Entropy &  &  & 5.09 & 4.65 & 3.98 & 4.25 & 4.30 & 4.01 & 4.39 & 3.64 & 3.59 & 4.25 & 5.28 & 4.33 \\
EAGLE3 &  &  & 5.16 & 4.91 & 4.13 & 4.32 & 4.42 & 4.10 & 4.50 & 3.79 & 3.83 & 4.38 & 5.46 & 4.48 \\
ELSI (m=3) &  &  & 5.23 & 4.91 & 4.15 & \textbf{4.40} & \textbf{4.49} & 4.18 & \textbf{4.61} & \textbf{3.87} & \textbf{4.02} & 4.51 & 5.54 & \textbf{4.57} \\
ELSI-C (m=3) &  &  & 5.19 & 4.91 & 4.14 & 4.32 & 4.46 & 4.16 & 4.49 & 3.83 & 3.92 & 4.47 & 5.52 & 4.52 \\
\midrule
EAGLE2 & \multirow{8}{*}{60} & \multirow{8}{*}{1.0} & 4.71 & 4.76 & 3.70 & 3.87 & 4.04 & 3.57 & 4.17 & 3.48 & 3.54 & 3.73 & 5.16 & 4.12 \\
ELSI (m=1) &  &  & 4.81 & 5.04 & 3.74 & 3.95 & 4.16 & 3.72 & 4.31 & 3.63 & 3.98 & 3.79 & 5.41 & 4.31 \\
\cmidrule{4-15}
ELSI (m=2) &  &  & 4.88 & 5.03 & \textbf{3.81} & 3.99 & 4.23 & 3.78 & 4.30 & 3.69 & 3.93 & 3.94 & \textbf{5.49} & 4.36 \\
ELSI-C (m=2) &  &  & 4.95 & \textbf{5.07} & 3.73 & 4.05 & 4.24 & 3.52 & 4.35 & 3.69 & \textbf{4.01} & 3.98 & 5.43 & 4.36 \\
\cmidrule{4-15}
Low-Entropy &  &  & 4.84 & 4.57 & 3.66 & 3.85 & 4.10 & 3.62 & 4.17 & 3.51 & 3.54 & 3.87 & 5.22 & 4.14 \\
EAGLE3 &  &  & \textbf{5.03} & 5.01 & 3.71 & 3.99 & 4.25 & 3.63 & 4.23 & 3.58 & 3.83 & 3.88 & 5.38 & 4.30 \\
ELSI (m=3) &  &  & 5.00 & 5.04 & 3.77 & \textbf{4.06} & \textbf{4.28} & 3.64 & \textbf{4.37} & \textbf{3.73} & 3.93 & \textbf{4.03} & 5.43 & \textbf{4.37} \\
ELSI-C (m=3) &  &  & 4.85 & 5.06 & 3.72 & 4.02 & 4.26 & \textbf{3.78} & 4.25 & 3.69 & 3.90 & 3.90 & 5.39 & 4.33 \\
\bottomrule
\end{tabular}
% }
\caption{Mean accepted tokens per forward pass, $\tau$, on SpecBench for Qwen3-4B-Instruct-2507, across draft-tree size (Nodes) and sampling temperature ($T$). Evaluation is performed with \textbf{EAGLE2, EAGLE3}, and \textbf{ELSI} and \textbf{ELSI-C} with layer-set sizes $m=1,2,3$. \textbf{Low-Entropy} is a negative control using layers $\{0,1,35\}$. M/R merges the math, reasoning, and math\_reasoning
categories, and \textbf{AVG} is the sample-weighted mean over all
categories. Bold marks the best value per block.}
\label{tab:qwen3-4b_block_efficiency_combined}
\end{table*}

\begin{table}
\centering
\small 
\setlength{\tabcolsep}{0.5mm}
% \resizebox{0.5\textwidth}{!}{
\begin{tabular}{lcccccccc}
\toprule
\textbf{Method} & \textbf{Nodes} & \textbf{T} & CODE & QA & ROLE. & WRIT. & M/R & \textbf{AVG} \\
\midrule
EAGLE2 & \multirow{7}{*}{30} & \multirow{7}{*}{0.0} & 5.36 & 4.09 & 4.15 & 4.41 & 4.89 & 4.40 \\
ELSI (m=1) &  &  & 5.51 & 4.19 & 4.25 & 4.54 & 5.15 & 4.58 \\
\cmidrule{4-9}
ELSI (m=2) &  &  & 5.51 & 4.22 & 4.26 & 4.57 & 5.14 & 4.58 \\
ELSI-C (m=2) &  &  & 5.49 & 4.19 & 4.13 & 4.54 & 5.08 & 4.53 \\
\cmidrule{4-9}
EAGLE3 &  &  & 5.54 & \textbf{4.24} & 4.25 & 4.62 & \textbf{5.22} & \textbf{4.64} \\
ELSI (m=3) &  &  & \textbf{5.55} & 4.22 & \textbf{4.30} & \textbf{4.62} & 5.14 & 4.59 \\
ELSI-C (m=3) &  &  & 5.47 & 4.15 & 4.24 & 4.51 & 5.10 & 4.52 \\
\midrule
EAGLE2 & \multirow{7}{*}{30} & \multirow{7}{*}{1.0} & 4.59 & 3.24 & 3.06 & 3.36 & 4.01 & 3.64 \\
ELSI (m=1) &  &  & 4.53 & 3.27 & \textbf{3.47} & 3.76 & 4.34 & 3.82 \\
\cmidrule{4-9}
ELSI (m=2) &  &  & 4.65 & 3.20 & 3.38 & 3.82 & 4.32 & 3.81 \\
ELSI-C (m=2) &  &  & 4.77 & 3.29 & 3.22 & 3.79 & 4.32 & 3.81 \\
\cmidrule{4-9}
EAGLE3 &  &  & 4.67 & 3.32 & 3.21 & \textbf{3.89} & \textbf{4.44} & \textbf{3.88} \\
ELSI (m=3) &  &  & \textbf{4.81} & \textbf{3.43} & 3.36 & 3.72 & 4.33 & 3.86 \\
ELSI-C (m=3) &  &  & 4.68 & 3.21 & 3.20 & 3.79 & 4.35 & 3.77 \\
\midrule
EAGLE2 & \multirow{7}{*}{60} & \multirow{7}{*}{0.0} & 5.58 & 4.36 & 4.48 & 4.61 & 5.24 & 4.66 \\
ELSI (m=1) &  &  & 5.73 & 4.42 & 4.57 & 4.85 & 5.47 & 4.84 \\
\cmidrule{4-9}
ELSI (m=2) &  &  & 5.70 & 4.49 & 4.54 & 4.84 & 5.42 & 4.83 \\
ELSI-C (m=2) &  &  & 5.67 & 4.44 & 4.46 & 4.81 & 5.42 & 4.79 \\
\cmidrule{4-9}
EAGLE3 &  &  & \textbf{5.75} & \textbf{4.51} & \textbf{4.60} & \textbf{4.88} & \textbf{5.51} & \textbf{4.89} \\
ELSI (m=3) &  &  & 5.75 & 4.47 & 4.59 & 4.85 & 5.45 & 4.84 \\
ELSI-C (m=3) &  &  & 5.66 & 4.39 & 4.52 & 4.80 & 5.37 & 4.78 \\
\midrule
EAGLE2 & \multirow{6}{*}{60} & \multirow{6}{*}{1.0} & 4.72 & 3.48 & 3.57 & \textbf{4.06} & 4.41 & 3.95 \\
ELSI (m=1) &  &  & \textbf{5.16} & 3.40 & 3.64 & 3.74 & 4.66 & 4.06 \\
\cmidrule{4-9}
ELSI (m=2) &  &  & 4.55 & 3.64 & 3.69 & 3.86 & 4.60 & 4.06 \\
ELSI-C (m=2) &  &  & 5.05 & 3.43 & 3.69 & 4.02 & 4.62 & 4.06 \\
\cmidrule{4-9}
EAGLE3 &  &  & 4.91 & \textbf{3.67} & 3.56 & 3.87 & 4.59 & \textbf{4.12} \\
ELSI (m=3) &  &  & 5.07 & 3.65 & \textbf{3.77} & 3.72 & 4.54 & 4.06 \\
ELSI-C (m=3) &  &  & 4.94 & 3.60 & 3.48 & 3.93 & \textbf{4.71} & 4.05 \\
\bottomrule
\end{tabular}
% }
\caption{Mean accepted tokens per forward pass, $\tau$, on SpecBench for Llama-3.1-8B-Instruct, across draft-tree size (Nodes) and sampling temperature ($T$). A subset of categories is shown for space. \textbf{AVG} is the sample-weighted mean over all SpecBench categories. M/R merges the math, reasoning, and math\_reasoning categories. Bold marks the best value per block.}
\label{tab:block_efficiency_llama_combined}
\end{table}

\section{Conclusion}
We studied multi-layer hidden-state selection for speculative decoding through the lens of information maximization and geometric diversity, moving beyond fixed, positional heuristics. Our analysis of decoder representations across \texttt{Qwen3-4B-Instruct} and \texttt{Llama-3.1-8B-Instruct} revealed consistent structural signals: a non-uniform entropy profile across layers (the activation wave), non-uniform entropy gains with layer concatenation, and CKA transition boundaries that indicate distinct representational compartments. Together, these observations motivate \emph{joint selection of layer sets} to maximize information and minimize redundancy, instead of relying on hand-crafted layer indices. Building on these signals, we introduced ELSI and ELSI-C as practical algorithms to select layer combinations for typical model sizes. We further enhanced these algorithms by introducing token-entropy filtering to reduce cross-prompt entropy variance, improving the stability of selection statistics. Empirically, we demonstrate (i) choice of layers matters more than the number of layers, and (ii) layer selection can result in sets that outperform EAGLE2 and EAGLE3 baselines. We further provide analysis of \emph{why} currently used layer combinations do perform well in practice, thus enhancing the interpretability of current speculative decoding techniques. Overall, this work provides a practical and interpretable framework for multi-layer hidden-state selection in speculative decoding. This framework is designed to be general, with the potential for future work that experiments with a broader variety of target models and draft-model conditioning strategies. Further, the selection of sets of layers may find application in a broader range of tasks beyond speculative decoding including embedding tasks and interpretability, and we encourage these as future directions for exploration.

% ============================================================
% REFERENCES
% ============================================================
\bibliographystyle{unsrt}

% IMPORTANT:
% Change this if your actual .bib filename is different.
\bibliography{aaai2027}

% ============================================================
% APPENDIX
% ============================================================
\appendix
\appendix
\section{Transition boundary comparison with and without token filtering}
\label{sec:appendix-cka-compare}

To isolate the effect of token filtering on representational compartment boundaries,
we compare default and token-filtered CKA transition plots side by side for both models.
% All four panels use the same transition-selection hyperparameters
% ($n_{\text{select}}=5$, $\text{min\_gap}=2$, $\text{min\_score\_frac}=0.01$),
% so differences are attributable to filtering rather than detector settings.

\begin{figure*}[t]
    \centering
    \begin{subfigure}[b]{0.48\linewidth}
        \includegraphics[width=\linewidth]{figures-v2/cka_transition_qwen3-4b_default.pdf}
        \caption{Qwen3-4B, default tokens}
    \end{subfigure}
    \hfill
    \begin{subfigure}[b]{0.48\linewidth}
        \includegraphics[width=\linewidth]{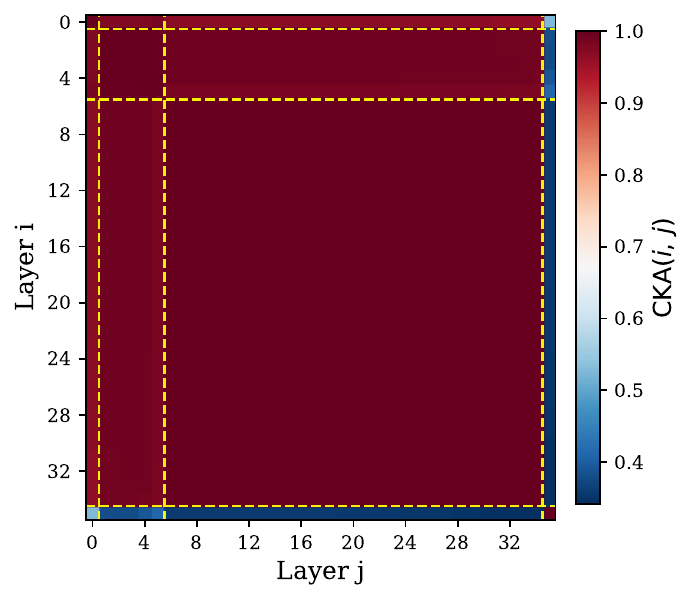}
        \caption{Qwen3-4B, token-filter ($\rho=0.30$)}
    \end{subfigure}

    \vspace{0.6em}

    \begin{subfigure}[b]{0.48\linewidth}
        \includegraphics[width=\linewidth]{figures-v2/cka_transition_llama-3.1-8b_default.pdf}
        \caption{Llama-3.1-8B, default tokens}
    \end{subfigure}
    \hfill
    \begin{subfigure}[b]{0.48\linewidth}
        \includegraphics[width=\linewidth]{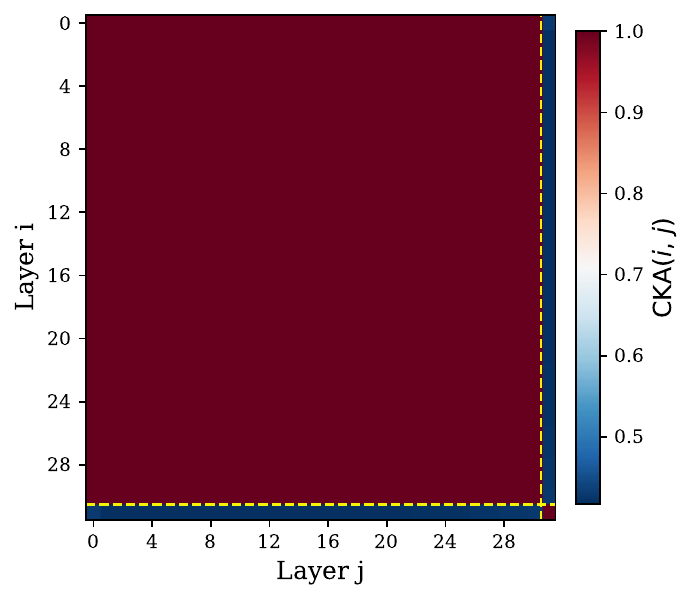}
        \caption{Llama-3.1-8B, token-filter ($\rho=0.30$)}
    \end{subfigure}
    \caption{With and without token-filter CKA transition comparison. Yellow dashed lines mark selected transition boundaries under matched detector settings.}
    \label{fig:appendix-cka-transition-compare}
\end{figure*}

% \begin{figure*}[t]
%     \centering
%     \begin{subfigure}[b]{0.48\linewidth}
%         \includegraphics[width=\linewidth]{figures-v2/topk_joint_entropy_s1_qwen3-4b.pdf}
%         \caption{Qwen3-4B-Instruct-2507}
%     \end{subfigure}
%     \hfill
%     \begin{subfigure}[b]{0.48\linewidth}
%         \includegraphics[width=\linewidth]{figures-v2/topk_joint_entropy_s1_llama-3.1-8b.pdf}
%         \caption{Llama-3.1-8B-Instruct}
%     \end{subfigure}
%     \caption{Entropy of layer pairs without Token-filtering. Top-$k$ layers by entropy are chosen as seeds and paired with the remaining layers $S_1([H_i;H_j])$.}
%     \label{fig:appendix-eg-topk-joint-compare}
% \end{figure*}

% \section{Global rank of different layers}
% We ran an exhasutive search over all layer combinations to compute global rank of ELSI compared to EAGLE2 and EAGLE3 and found that given ELSI already has higher joint entropy than EAGLE2/3, the layers selected also has way better global rank when an exhaustive run is done over all thousands of combination as shown in Table~\ref{tab:entropy-stacked}
% \input{tables/layer_rank_trim}

\section{ELSI Beam Analysis}
\begin{figure*}[t]
    \centering
    \includegraphics[width=0.9\linewidth]{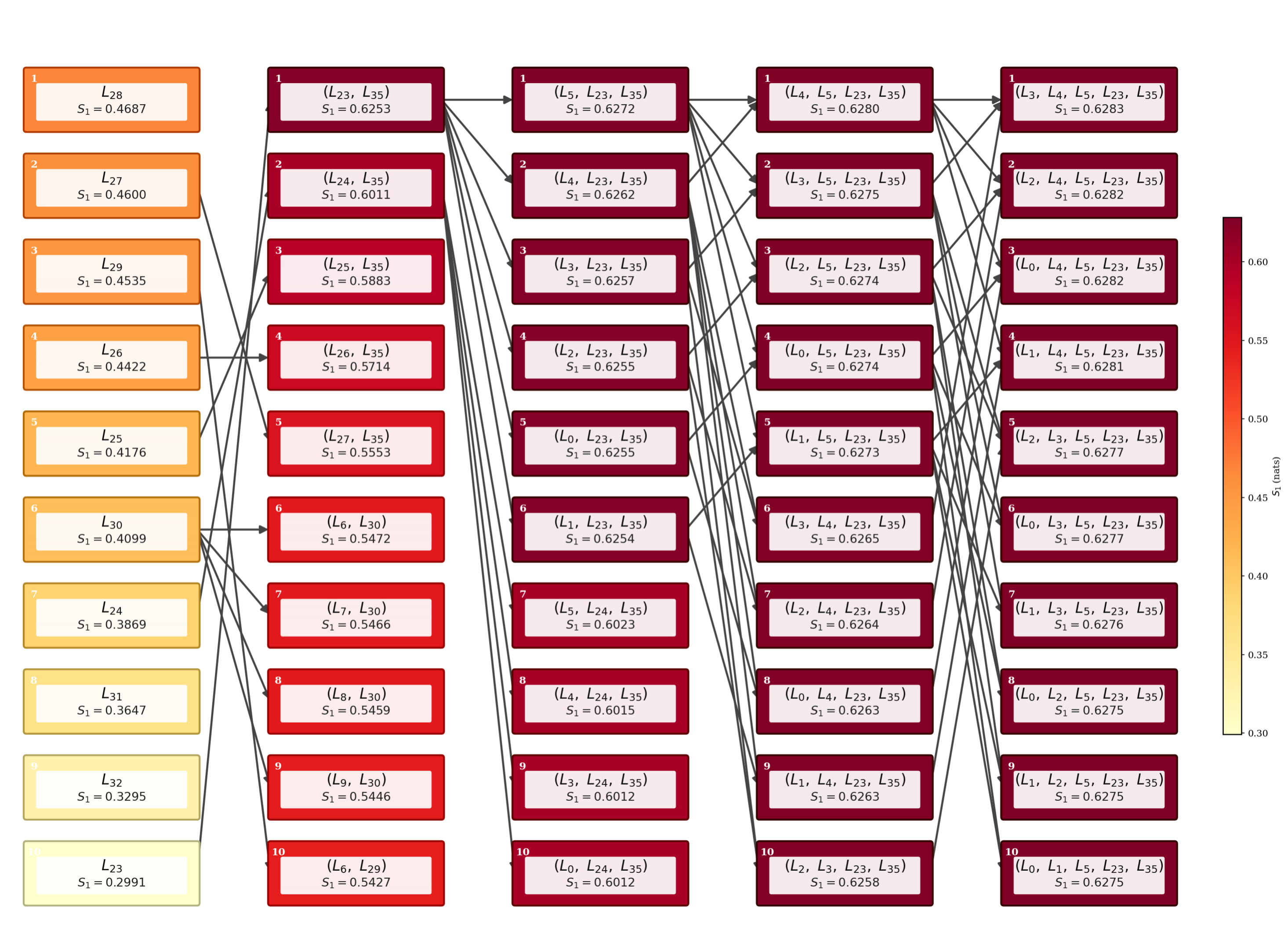}
    \caption{Top-M entropy beam frontier for \texttt{Qwen3-4B-IT}}
    \label{fig:beam-baseline}
\end{figure*}

We show the entire beam tree using ELSI, expanding based on high joint entropy of hidden-states post token-filtering in Fig~\ref{fig:beam-baseline}. The top-1 at each depth of the beam are layers picked for best single, pair, and triple layers respectively.

% ============================================================
% END DOCUMENT
% ============================================================
\end{document}